\documentclass[preprint,12pt]{article}

\usepackage{style}
\usepackage{amssymb}
\usepackage{soul}
\usepackage{booktabs}
\usepackage{amsmath}
\usepackage{cleveref}
\usepackage[T1]{fontenc}
\usepackage[utf8]{inputenc}
\usepackage{lmodern}
\usepackage{acro}
\usepackage{multirow}
\usepackage{url}
\usepackage{tabularx}
\DeclareAcronym{art}{short = ART, long  = approximate randomization testing, cite = Riezler2005pitfalls}
\DeclareAcronym{beer}{short = BEER, long = BEtter Evaluation as Ranking, cite = stanojevic2014beer}
\DeclareAcronym{bleu}{short = BLEU, long  = bilingual evaluation understudy, cite = papineni2002bleu}
\DeclareAcronym{bwer}{short = bWER, long = bag-of-words WER, cite = Vidal2023bwer}
\DeclareAcronym{bpe}{short = BPE, long  = byte pair encoding, cite = Gage94}
\DeclareAcronym{cbmt}{short = CBMT, long  = character-based machine translation}
\DeclareAcronym{cbsmt}{short = CBSMT, long  = character-based statistical machine translation}
\DeclareAcronym{cbnmt}{short = CBNMT, long  = character-based neural machine translation}
\DeclareAcronym{chrf}{short = chrF, long = F-score based on character n-grams, cite = popovic2015chrf}
\DeclareAcronym{cm}{short = CM, long  = confidence measures}
\DeclareAcronym{cer}{short = CER, long  = character error rate}
\DeclareAcronym{fda}{short = FDA, long  = feature decay algorithm, cite = Biccici15}
\DeclareAcronym{hplt}{short = HPLT, long = High Performance Language Technologies, cite = gilbert2024new}
\DeclareAcronym{htr}{short = HTR, long  = handwritten text recognition}
\DeclareAcronym{il}{short = IL, long  = incremental learning}
\DeclareAcronym{imt}{short = IMT, long  = interactive machine translation}
\DeclareAcronym{inmt}{short = INMT, long  = interactive neural machine translation}
\DeclareAcronym{ismt}{short = ISMT, long  = interactive statistical machine translation}
\DeclareAcronym{ksr}{short = KSR, long  = key stroke rate}
\DeclareAcronym{llm}{short = LLM, long  = large language model}
\DeclareAcronym{lstm}{short = LSTM, long  = long short-term memory, cite = Hochreiter97}
\DeclareAcronym{mar}{short = MAR, long  = mouse action rate}
\DeclareAcronym{mt}{short = MT, long  = machine translation}
\DeclareAcronym{nlp}{short = NLP, long  = natural language processing}
\DeclareAcronym{nmt}{short = NMT, long  = neural machine translation}
\DeclareAcronym{ocr}{short = OCR, long = optical character recognition}
\DeclareAcronym{ood}{short = OOD, long = out-of-distribution, cite = hendrycks2016ood}
\DeclareAcronym{pe}{short = PE, long  = post-editing}
\DeclareAcronym{rbmt}{short = RBMT, long  = rule-based machine translation}
\DeclareAcronym{relu}{short = ReLU, long  = rectified linear unit}
\DeclareAcronym{rlhf}{short = RLHF, long = reinforcement learning with human feedback, cite = lambert2025rlhf}
\DeclareAcronym{rnn}{short = RNN, long  = recurrent neural network, cite = {Hochreiter97}}
\DeclareAcronym{sgd}{short = SGD, long  = stochastic gradient descend}
\DeclareAcronym{smt}{short = SMT, long  = statistical machine translation}
\DeclareAcronym{ter}{short = TER, long  = translation error rate, cite = Snover2006ter}
\DeclareAcronym{wer}{short = WER, long  = word error rate}
\DeclareAcronym{wsr}{short = WSR, long  = word stroke rate}
\DeclareAcronym{xml}{short = XML, long  = eXtensible Markup Language}
\usepackage{xcolor}
\usepackage{array}
\usepackage{authblk}
\usepackage{lineno}
\usepackage{graphicx}
\usepackage{algorithm}
\usepackage{algpseudocode}

\title{DEEP: Docker-based Execution and Evaluation Platform}
\begin{document}

\author[1]{Sergio G\'omez Gonz\'alez$^\dagger$}
\author[1,2]{Miguel Domingo}
\author[1,2]{Francisco Casacuberta}

\affil[1]{PRHLT Research Center - Universitat Polit{\`e}cnica de Val{\`e}ncia}
\affil[2]{ValgrAI - Valencian Graduate School and Research Network for Artificial Intelligence}

\email{$^\dagger$sgomgon@prhlt.upv.es}

\maketitle

\begin{abstract}
Comparative evaluation of several systems is a recurrent task in researching. It is a key step before deciding which system to use for our work, or, once our research has been conducted, to demonstrate the potential of the resulting model. Furthermore, it is the main task of competitive, public challenges evaluation. Our proposed software (DEEP) automates both the execution and scoring of machine translation and optical character recognition models.  Furthermore, it is easily extensible to other tasks.
\emph{DEEP} is prepared to receive dockerized systems, run  them (extracting information at that same time), and assess hypothesis against some references. With this approach, evaluators can achieve a better understanding of the performance of each model. Moreover, the software uses a clustering algorithm based on a statistical analysis of the significance of the results yielded by each model, according to the evaluation metrics. As a result, evaluators are able to identify clusters of performance among the swarm of proposals and have a better understanding of the significance of their differences. Additionally, we offer a visualization web-app to ensure that the results can be adequately understood and interpreted. Finally, we present an exemplary case of use of \emph{DEEP}.

\end{abstract}

\begin{figure}
    \centering
    \includegraphics[width=0.75\linewidth]{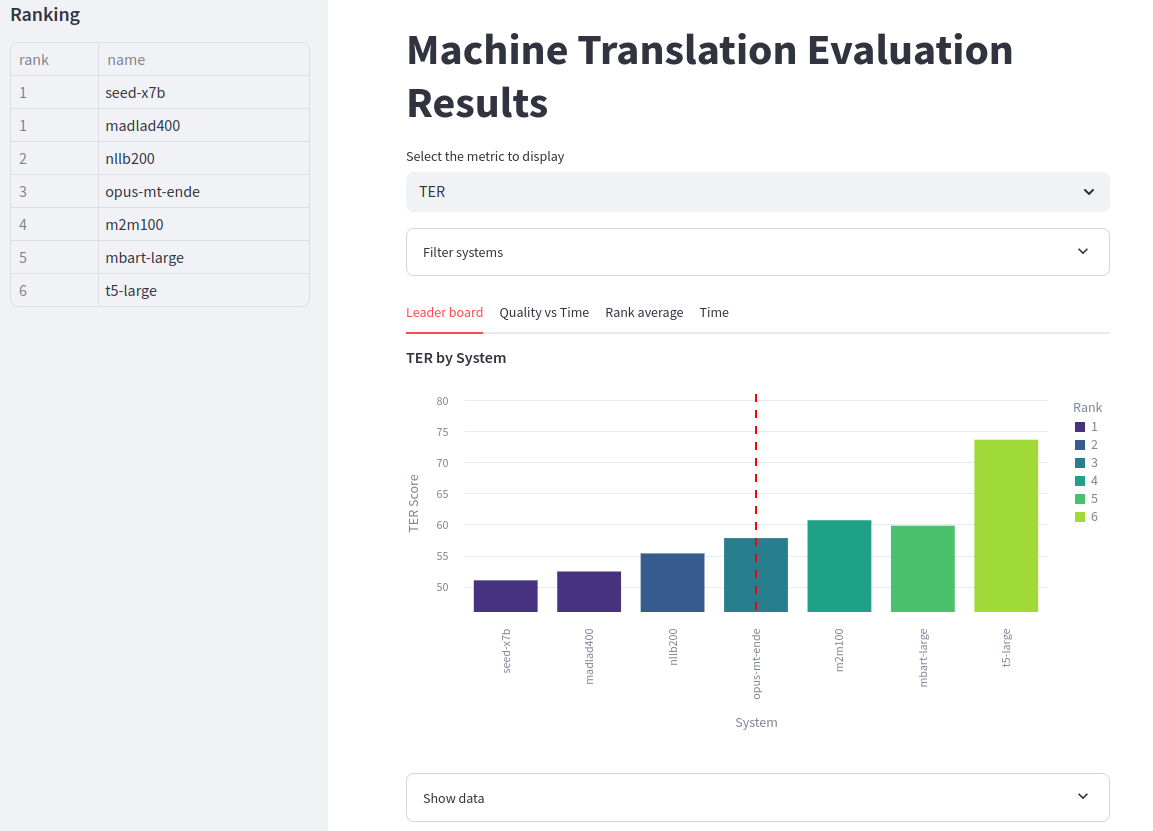}
    \caption{Visual interface of DEEP with the results of an example.}
    \label{fig:overview}
\end{figure}
\section{Introduction}
When organizing a competitive machine learning task, a recurrent problem is evaluating the predictions of different systems with different resource needs. In the classical approach, participants receive the source part of the test sets and are asked to send their hypotheses. However, with this approach, evaluators cannot analyze the amount of resources that each submitted system takes to obtain its hypothesis. 

Other approaches ask participants to submit their systems and, then, evaluators run the test sets over them. This approach ensures even more opacity to hide the test set since participants no longer have access to it. In this work, we present \emph{\textbf{D}ocker-based \textbf{E}xecution and \textbf{E}valuation \textbf{P}latform}\footnote{\url{github.com/sergiogg-ops/deep}} (DEEP). With it, the execution of each proposal can be monitored and measured to extract even more information. We believe that our proposal can serve as a reference from which official competitions and individual researchers can extract code for their specific applications.

The software that we propose has been designed for \ac{mt} and \ac{ocr} purposes. However, its architecture is as modular as possible to easily adapt to any other machine learning tasks. It is composed of two phases: evaluation and visualization. The first one is divided into two steps as well: the execution of the proposals and the evaluation of the hypotheses. The first step is optional and can be avoided if the systems to evaluate are not available. In that case, the information about the execution will not be available either for the analysis. The result of this phase is a file that contains the scores for different metrics of the various models and information about the execution. The evaluators can visualize these results as many times as required in the second phase.

The rest of the article is structured as follows: in Section 2 we present an introduction to the current state-of-the-art, in Section 3 we describe the description of the application, Section 4 contains the list of metrics that \emph{DEEP} allows currently, in Section 5 we present a case of use of \emph{DEEP} as an example and, finally, we will conduct some conclusions and future work in Section 6.
\section{State of the art}
Current objective evaluation and analysis is supported by metrics that measure different aspects of similarity with respects to a reference \cite{kocmi2023findings,kocmi2024wmt}. To fulfill a complete analysis from an evaluation process, one needs to look to several supplementary metrics. For this purpose, many metrics have been proposed \cite{Snover2006ter,papineni2002bleu,Vidal2023bwer,sparck1972fscore}. 

Advances in generative machine learning have brought new challenges. There are voices that claim that classical statistical metric scores are not enough \cite{gao2025llm}. They propose metrics based on neural models like \ac{llm}. Some of them are based on the comparison between the embeddings of the hypothesis and reference sequences, such as \emph{BertScore} \cite{zhang2019bertscore} or \emph{COMET} \cite{rei2020comet}. It has also been proposed to use \ac{llm}s with \ac{rlhf} to align their scores with human criteria \cite{gao2025llm}. However, this last proposal has not been introduced as a commonly used evaluation protocol.

The main flaw of these evaluation systems is that it is hard to know the reasons why they produce a certain score. Both LLM-evaluators and the systems they evaluate are based on the Transformer architecture \cite{vaswani2017attention}. These kinds of systems, such as neural models, are difficult to interpret beyond determining whether they are wrong or right \cite{ali2022xai}. In order to speculate on how the models would perform in a real scenario, it is important to determine the causes of their successes and failures. That is precisely the goal of the explainability branch of machine learning research \cite{ali2022xai,nguyen2023inspecting,kashefi2023explainability}.

In the evaluation process, it is also important to consider the data on which the models were trained and the data on which they are evaluated. Performance in data similar to the training set is likely to be the same as that observed during the training phase. Whereas the behaviour when presented with data of different properties is more unpredictable \cite{li2025probing}. This is the problem of the \ac{ood} data, which presents a good evaluation method to test the generalizability of machine learning models \cite{kim2024investigation}.

Further from the simple error aggregation rates it is important to evaluate systems in a holistic manner \cite{deldjoo2025toward,haag2024development}. This means to analyze the robustness, fairness, efficiency and user impact of the evaluated model. Some fields such as \ac{mt}, summarization or others require an estimation of the quality on the answer. To perform a complete evaluations of systems like those it is important to provide a study on how humans perceive the hypotheses of the systems \cite{freitag2021experts}. Often, machine learning models are not designed to work alone but as support to the work of a human. These human-in-the-loop \cite{wu2022survey} schemes require the models to adjust well to human way of working. Therefore, for these schemes is vital to receive human feedback to adequately evaluate different systems.
\section{How DEEP works}
\emph{DEEP} is a pipeline consisting of three phases: execution, evaluation and visualization. If the evaluators only have access to the hypotheses, the first step (execution) can be avoided. The information that would be extracted on that step will not be available but the rest of the functions will stay the same.
\subsection{Docker-based execution step}
Different tasks require data of different nature. We selected \ac{mt} and \ac{ocr} to showcase that our software successfully works, independently of the data being text or images. In our case, the \ac{mt} dataset was specified in two \emph{xml} files: source and target. Meanwhile, the images for \ac{ocr} are stored in a directory, and the transcripts are in several plain text files with the same names. Note that you can adapt the format of the data by simply changing the input function, maintaining the rest of the functionality.

Our application requires the systems to be packaged in dockerfiles. With them, the software builds the image and, then, it executes the system. Depending on the task, the source test set is fed to the docker as a file or a directory in the `data' sub-directory. The hypotheses are gathered after the execution in the `predictions' file or folder of the same sub-directory. The systems are required to contain the needed functionality to match this behaviour. When all systems have been executed, all images and waste are removed to avoid excessive disc usage.

The predictions made by the proposals are stored in another directory identified by the names of the dockerized systems. Furthermore, the execution time is fed into the next step to be stored as information in the results file. 
\begin{algorithm}
\caption{Algorithm to compute the clusters among all of the systems according to the statistical significance of their differences \cite{casacuberta2024findings}.}
\label{alg:clustering}
\begin{algorithmic}
\State $Systems \gets [System1, System2, ..., SystemN]$
    \State $Systems \gets Sort(Systems)$
    \State $clusters \gets [[Systems[0]]]$
    \State $i \gets 1$
    \While{$i < N$}
    \State $prev \gets Systems[i-1]$
    \State $this \gets Systems[i]$
    \State $i \gets i+1$
    \If{$art(prev,this,trials=10000,p=0.05)$}
    \State $clusters.add([this])$
    \Else
    \State $clusters[-1].add(this)$
    \EndIf
    \EndWhile
\end{algorithmic}
\end{algorithm}

\subsection{Evaluation step}
As a first step, the predictions obtained in the previous phase are loaded. If the dataset contains multiple files, they are aggregated to evaluate it as one larger dataset. Along with them, the references are loaded in the same way, and the evaluation process can start. After that, metrics are computed, systems are clustered, and finally, a file containing this information is created.

\subsubsection{Score measurement}
All systems are evaluated using the selected metrics. For each metric, both the general and sample scores are obtained. While the first is shown to the user, the latter is useful for computing statistical tests at a later step.

Each metric has an assigned function to be computed, which accepts the same input: a list of hypotheses and a list of references. So far, our system can evaluate \ac{bleu} and \ac{ter} for \ac{mt}, and \ac{wer} and \ac{bwer} for \ac{ocr}. If the user wanted to include any other metric, they would need to accept the same input and offer a \emph{float} output score as return value.

\subsubsection{Clustering}
Scores alone are not enough to determine which system performs better. It is important to determine whether the differences between them are actually significant \cite{barrault2020findings}. To this end, the most common approach is to compute a statistical significance test \cite{dietterich1998approximate,ulmer2022deep,rainio2024evaluation}. \Ac{art} has demonstrated to be very useful in fields of \ac{nlp} such as \ac{mt}. Furthermore, it is a metric agnostic test, convenient for our modular-orientated app. Our software also classifies the systems into different clusters  based on the statistical significance of their differences \cite{casacuberta2024findings,barrault2020findings} obtained with \ac{art}. 

For this purpose, systems are first sorted by the main metric. Then, \ac{art} with $10000$ trials and a p-value of $0.05$ is computed between the adjacent proposals in sorting order. Finally, from the best to worst scoring system, if \ac{art} indicates that the differences between one system and the previous are not significantly different, then the system is added to the same cluster as the previous one. Otherwise, a new cluster is created and that system is added to it \cite{casacuberta2024findings}.

\subsection{Visualization}
We also created a tool to visualize the results obtained from the aforementioned evaluation process. This last part of the pipeline has been created using the library \emph{Streamlit}\footnote{\url{streamlit.io}}. Following our proposal of the \ac{mt}--\ac{ocr} mixed pipeline, the title is set to one option or another. The available metrics are those that were evaluated in the previous evaluation process and will depend on the evaluation task.
\begin{figure}
    \centering
    \includegraphics[width=0.75\linewidth]{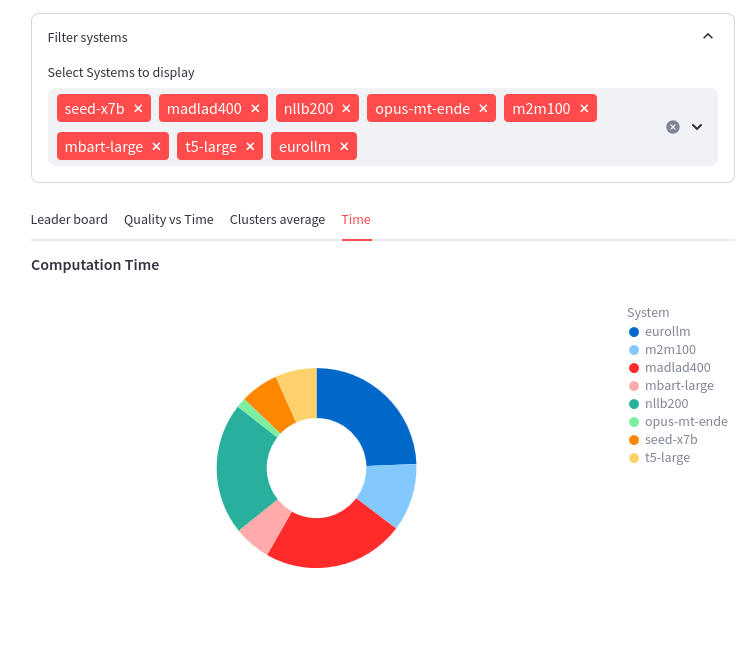}
    \caption{Chart of the computation time needed by the proposals of a demo with DEEP.}
    \label{fig:pie_chart}
\end{figure}

The visual interface allows the user to change the metric that is displayed at any time. For that purpose, a selector is placed at the top of the screen (\cref{fig:overview}). As another filter, the user can select which proposals are displayed to focus on the analysis of just some of them. Furthermore, a table is displayed at the lateral bar, sorting systems according to the selected metric and ranking them according to the clustering algorithm described at the previous section.

\begin{figure}
    \centering
    \includegraphics[width=0.75\linewidth]{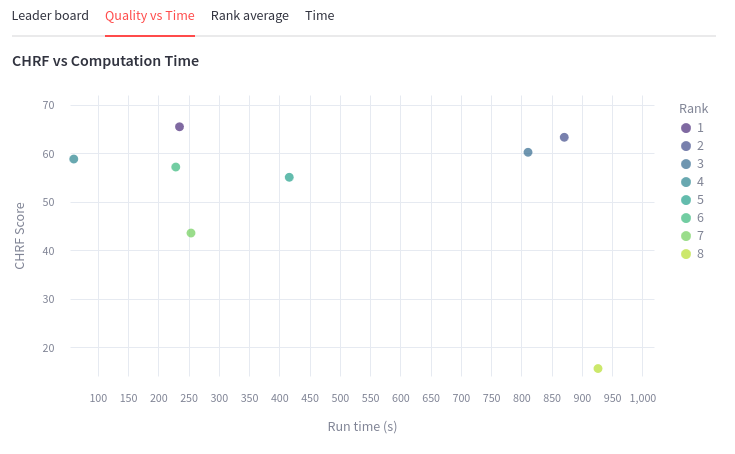}
    \caption{Chart of time vs quality of hypothesis of a demo with DEEP.}
    \label{fig:scatterplot}
\end{figure}

\begin{figure}
    \centering
    \includegraphics[width=0.75\linewidth]{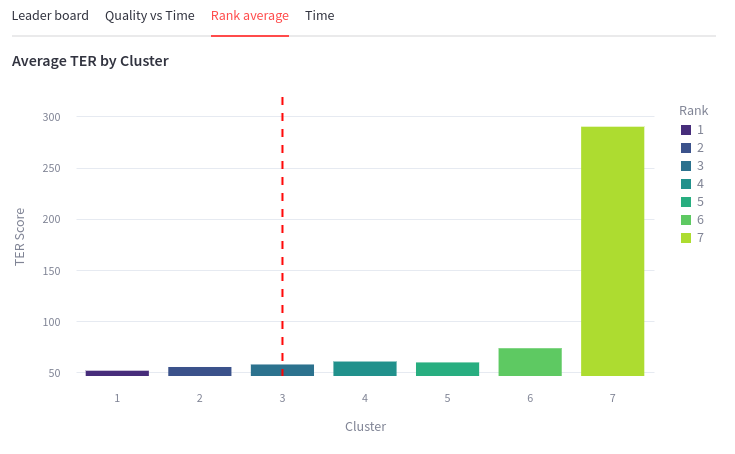}
    \caption{Chart of average by cluster of a demo with DEEP.}
    \label{fig:cluster}
\end{figure}

The most informative figure is displayed in the center of the screen. In that place, the user can visually check a chart for the scores of all systems. Furthermore, by changing the tab of the chart, different display options can be selected:
\begin{description}
    \item [Basic:] a bar chart displaying the score for each system, ordered from best to worst. Baselines are highlighted with a red dashed line.  \cref{fig:overview} displays an example of this functionality.
    \item [Time:] an arc chart displaying the computation time needed by each system. A visual example is available at \cref{fig:pie_chart}.
    \item [Overall:] a scatter plot displaying the score for each system against the time it took to obtain its hypotheses. \cref{fig:scatterplot} presents a demo of this chart.
    \item [Clusters:] a bar chart displaying the average score of all the proposals from each cluster. Clusters that contain a baseline are highlighted with a red dashed line. \cref{fig:cluster} displays an example of this feature.
\end{description}
Finally, at the bottom of the interface (\cref{fig:overview}), the software allows the user to visualize the data contained in the results file. It is adequately displayed in a table in an expander, as shown in \cref{fig:data}. This table can be exported to \emph{CSV}, \emph{LaTeX}, \emph{JSON} or \emph{HTML}.

\section{Included metrics} \label{sec:metrics}
The current version of our software implements the following six metrics, which can be easily extended with any other according to different needs.

\begin{description}
    \item[BLEU] \cite{papineni2002bleu} is an \ac{mt} metric that computes the geometric average of the modified \emph{n-gram} precision, multiplied by a brevity factor that penalizes short sentences. To ensure that our scores are consistent with other studies, we made use of \emph{sacreBLEU}\footnote{\url{github.com/mjpost/sacrebleu}} \cite{Post2018bleu}.
    \item[TER] \cite{Snover2006ter} is an \ac{mt} metric that computes the number of word edit operations (insertion, substitution, deletion and swapping), normalized by the number of words in the reference. To ensure that our scores are consistent with those of any other studies, we made use of \emph{sacreBLEU} \cite{Post2018bleu}.
    \item[chrF] \cite{popovic2015chrf} is an \ac{mt} metric that applies statistics after the common F-score to assess the similarity between sentences using \emph{n-grams}. To ensure that our scores are consistent with those of any other studies, made use of \emph{sacreBLEU} \cite{Post2018bleu}.
    \item[BEER] \cite{stanojevic2014beer} is an \ac{mt} metric that combines adequacy features with ordering features into a linear model. This model is trained using learning-to-rank techniques to align with human judgments of translation quality.
\end{description}

\section{Case of use}
To showcase the functioning of DEEP, we have prepared a demo trial with a case of use. As an example, we will perform an \ac{mt} study on the current best performing models that are open-source. We will use DEEP as our only tool for execution and analysis to demonstrate its potential.

\subsection{Experimental setup}
We have selected a set of state-of-the-art models for the purpose of testing them using \emph{DEEP} on the test set of the \emph{News} shared task from \emph{WMT20} \cite{barrault2020findings}. Said models can be downloaded and used with \emph{HuggingFace}'s \emph{transformers}\footnote{\url{huggingface.co/docs/transformers/es/index}}:

\begin{figure}
    \centering
    \includegraphics[width=0.75\linewidth]{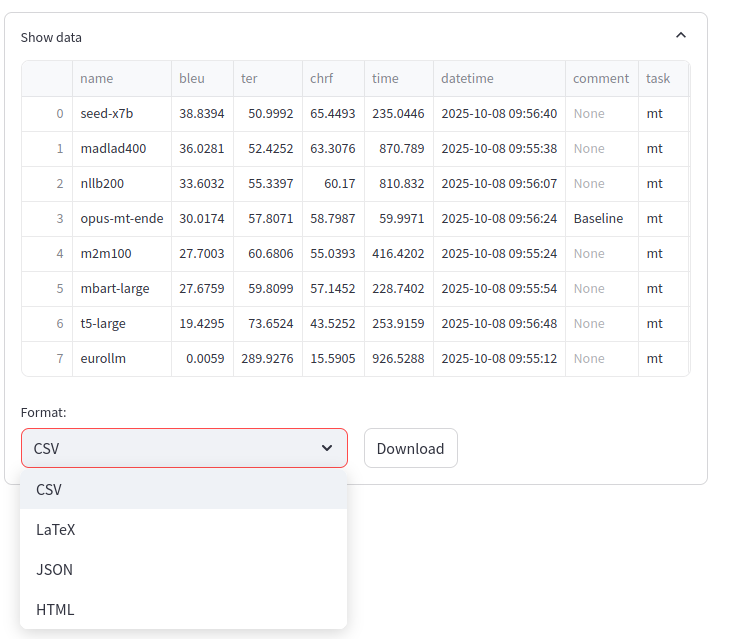}
    \caption{Display of the evaluation data and download options.}
    \label{fig:data}
\end{figure}

\begin{description}
    \item[EuroLLM\footnotemark]\footnotetext{\url{huggingface.co/utter-project/EuroLLM-1.7B}} \cite{martins2024eurollm}: multilingual \ac{llm} trained with the support of the European Union. It is not a specific translation model, but its multilingual nature makes it suitable for the task. We have used the version with 1.7 billion parameters.
    \item[M2M-100\footnotemark]\footnotetext{\url{huggingface.co/facebook/m2m100_1.2B}} \cite{fan2021m2m}: the multilingual \ac{mt} model developed by the former \emph{Facebook Research} (now \emph{Meta AI}) team is one of the first widely multilingual models. We have used the version with 1.2 billion parameters.
    \item[MADLAD-400\footnotemark]\footnotetext{\url{huggingface.co/google/madlad400-3b-mt}} \cite{kudugunta2023madlad}: multilingual translation \ac{llm} trained by \emph{Google Deepmind} and made available under the \emph{Apache-2.0} license. We have used the version with 3 billion parameters.
    \item[Opus-MT English-to-German\footnotemark]\footnotetext{\url{huggingface.co/Helsinki-NLP/opus-mt-en-de}} \cite{Tiedemann2020opusmt}: trained with the \emph{Marian-NMT}\footnote{\url{marian-nmt.github.io}} framework, it is part of a group of direction-specific \ac{mt} models. They were trained in the \emph{Opus-MT} project that produced both new neural \ac{mt} models and datasets.
    \item[mBART\footnotemark]\footnotetext{\url{https://huggingface.co/facebook/mbart-large-50-many-to-many-mmt}} \cite{liu2020mbart}: the model trained by \emph{FAIR} (now \emph{Meta AI}) is one of the first multilingual translation models. We have used the large version with 611 million parameters.
    \item[NLLB-200\footnotemark]\footnotetext{\url{huggingface.co/facebook/nllb-200-3.3B}} \cite{costa2022nllb}: multilingual \ac{mt} model obtained from the \emph{No Language Left Behind} project by \emph{Meta AI}. We have used the version with 3.3 billion parameters.
    \item[Seed\footnotemark]\footnotetext{\url{huggingface.co/ByteDance-Seed/Seed-X-PPO-7B}} \cite{cheng2025seed}: developed by the \emph{Byte Dance} lab, it constitutes an open-source \ac{mt} model whose authors claim it is competitive with closed-source models. However, its language span is restricted to ``just'' 28 languages, with a size of 7 billion parameters.
    \item[T5\footnotemark]\footnotetext{\url{huggingface.co/google-t5/t5-large}} \cite{Xue2021mt5}: the model developed by \emph{Google Research} is one of the first generalist \ac{nlp} models. It is not translation specific, but it can be conditioned by a prompt to develop this task. We have used the large version with 738 million parameters.
\end{description}

We have dockerized each system with the corresponding behaviour to match the restrictions of the \emph{DEEP} framework.
As stated above, we selected the test set from WMT20's News shared task\cite{barrault2020findings}. More precisely, we selected the English-to-German task with 1418 sentences. We performed the experiment on a machine equipped with an \emph{Nvidia Quadro RTX 8000} with 48 GB of \emph{VRAM}. 

\subsection{Analysis}
After obtaining the results from the evaluation phase, we easily converted them to \cref{tab:results} using DEEP's visualization tool. Furthermore, we instantly identified the three models that produced the best hypotheses: \emph{Seed}, \emph{MADLAD-400} and \emph{NLLB-200} (in that order). The tool orders the model according to the scores obtained with the best metric, so we can rapidly rank the models. Thus, the actual order from best-to-worst is displayed at \cref{tab:results} and at \cref{fig:overview}. Sometimes, the order is not the same for all metrics. Actually, in our example, the order for the \ac{chrf} metric is inverted for \emph{M2M-100} and \emph{mBART} models.

We were able to check the execution time of each model using the chart from \cref{fig:pie_chart}. Thus, we can conclude that the fastest models are \emph{Opus-MT}, \emph{mBART} and \emph{Seed}. With this type of pie chart, one cannot know the actual value of each slice. However, in our interface, placing the cursor over each area reveals the exact computation time. This behaviour is replicated for any chart of the visualization tool.

One of the most interesting conclusions of our experiment is visualized in the plot of \cref{fig:scatterplot}. The model \emph{Seed}{\textemdash}besides being the one with the best performance{\textemdash}is also one of the fastest. It represents a clear improvement over the state of the art, as it obtains better translations faster than previous open-source models. Obviously, to clearly state such a strong affirmation, a larger study is needed. However, that is out of the scope of our demo. Moreover, from the plot, we can separate three groups:
\begin{itemize}
    \item \textbf{Balanced models}: this group is made up of \ac{mt} models that balance performance with speed. It contains both the best translator (\emph{Seed}) and the fastest (\emph{Opus-MT}), along with some others: \emph{mBART}, \emph{T5} and \emph{M2M-100}.
    \item \textbf{Heavy models}: this group contains models that scored well in the metrics at the expense of computation time. It contains the \emph{MADLAD-400} and \emph{NLLB-200} models.
    \item \textbf{Improvement zone}: this group is composed of a single model: \emph{EuroLLM}. It is a model that is rather slow and does not score well on the metrics. However, with further iterations, different prompting or other datasets, it could achieve better performance.
\end{itemize}

For the metrics \ac{bleu} and \ac{chrf}, all differences are statistically significant. That means that each model has its own cluster, and no cluster groups several models. However, for the \ac{ter} metric, \emph{Seed} and \emph{MADLAD-400} share the same cluster. Therefore, for the \ac{ter} metric there isn't a single model that yields the best results.

\begin{table}
    \centering
    \resizebox{0.5\textwidth}{!}{
    \begin{tabular}{lrrrr}
        \toprule
        System & BLEU & TER & chrF & Time (s) \\
        \midrule
        Seed-x7b & $38.84$ & $51.00$ & $65.45$ & $236.28$ \\
        MADLAD-400 & $36.03$ & $52.43$ & $63.31$ & $870.62$ \\
        NLLB-200 & $33.60$ & $55.34$ & $60.17$ & $774.95$ \\
        OPUS-MT & $30.02$ & $57.81$ & $58.80$ & $57.60$ \\
        M2M-100 & $27.70$ & $60.68$ & $55.04$ & $417.13$ \\
        mBART & $27.68$ & $59.81$ & $57.15$ & $228.66$ \\
        T5-large & $19.43$ & $73.65$ & $43.53$ & $252.56$ \\
        EuroLLM & $0.85$ & $279.03$ & $22.11$ & $865.11$ \\
        \bottomrule
        \end{tabular}
        }
    \caption{Results of the demo of DEEP.}
    \label{tab:results}
\end{table}

\section{An extension of DEEP}
In the previous section, we exposed a case of use of our tool for an \ac{mt} task. Nevertheless, \emph{DEEP} also works for \ac{ocr} as stated before. The input images should be contained into a single directory, and the containers should read the images and produce the transcripts. For the evaluation of this task we have implemented the following well-known metrics:

\begin{description}
    \item[WER] \cite{morris2004and} is an \ac{ocr} metric that is directly derived from the \emph{Levenshtein} or edit distance at a word level.
    \item[Bag-of-words WER] \cite{Vidal2023bwer} is an \ac{ocr} metric that computes \ac{wer} to the hypotheses as if they were a bag-of-words instead of an ordered list of words. To ensure consistency, we used the implementation by \citet{Vidal2023bwer}\footnote{\url{github.com/PRHLT/E2EHTREval}}.
\end{description}

Nonetheless, the metrics presented in \cref{sec:metrics} can also be used for this task. The rest of the features of the pipeline work in the same way. The clustering algorithm is applied to the results as well, and the visualization tool can be used at the full of its potential. 
\section{Conclusions}
In this work, we proposed an evaluation framework with an implementation that is ready to use and easy to extend: DEEP. It automates the execution of different \ac{nlp} systems to produce hypotheses over one or more test sets. The only requirement is that the systems are enclosed in a Docker container that produces the hypotheses of the models for the input. After that, the system evaluates the hypotheses with a metric chosen by the use{\textemdash}our current version comes with six different metrics, which can be easily extended to meet users' needs. This information is combined with the computation time required to obtain the hypotheses and is stored in a results file. Furthermore, DEEP integrates a visualization web application that allows the user to inspect these results. This application contains several tools and charts to help understand and interpret the performance of the different models.

In order to demonstrate the usefulness of \emph{DEEP}, we have performed an exemplary experiment. A set of 8 state-of-the-art models was selected and tested on WMT20's test partition. As a result, we were able to conduct a complete comparative study, the only flaws of which can be attributed to the size and scope of the test set.

In general, we have designed an end-to-end evaluation pipeline for \ac{mt} and \ac{ocr}. Since we are making it open-source, the user can adapt it to their needs. Nonetheless, we plan to expand the functionality of DEEP to a wider set of tasks. This will mean allowing new types of data into the pipeline and including the appropriate metrics for them, among others.

\section{Acknowledgements}
This work received funding from \emph{ValgrAI}; \emph{Generalitat Valenciana}; \emph{European NextGeneration EU/PRTR} (project FAKEnHATE-PdC; PDC2022-133118-I00); and \emph{MCIN/AEI} and \emph{ERDF/EU} (project LLEER; PID2021-124719OB-I00).

\bibliographystyle{apalike} 
\bibliography{imt}

@inproceedings{papineni2002bleu,
  title         = {{BLEU}: a method for automatic evaluation of machine translation},
  author        = {Papineni, Kishore and Roukos, Salim and Ward, Todd and Zhu, Wei-Jing},
  booktitle     = {Proceedings of the 40th annual meeting of the Association for Computational Linguistics},
  pages         = {311--318},
  year          = {2002}
}

@article{vaswani2017attention,
  title         = {Attention is all you need},
  author        = {Vaswani, Ashish and Shazeer, Noam and Parmar, Niki and Uszkoreit, Jakob and Jones, Llion and Gomez, Aidan N and Kaiser, {\L}ukasz and Polosukhin, Illia},
  journal       = {Advances in neural information processing systems},
  volume        = {30},
  year          = {2017}
}

@inproceedings{kocmi2023findings,
  author        = {Kocmi, Tom and Avramidis, Eleftherios and Bawden, Rachel and Bojar, Ond{\v{r}}ej and Dvorkovich, Anton and Federmann, Christian and Fishel, Mark and Freitag, Markus and Gowda, Thamme and Grundkiewicz, Roman and others},
  title         = {Findings of the 2023 Conference on Machine Translation (WMT23): LLMs Are Here but Not Quite There Yet},
  booktitle     = {Proceedings of the Eighth Conference on Machine Translation},
  year          = {2023},
  pages         = {1--42},
}

@inproceedings{Post2018bleu,
    title			=	{A Call for Clarity in Reporting BLEU Scores},
    author			=	{Post, Matt},
    booktitle		=	{Proceedings of the Third Conference on Machine Translation},
    pages			=	{186--191},
    year			=	{2018}
}

@inproceedings{Snover2006ter,
    title       	= 	{A Study of Translation Edit Rate with Targeted Human Annotation},
    author      	= 	{Snover, Matthew and Dorr, Bonnie and Schwartz, Richard and Micciulla, Linnea and Makhoul, John},
    booktitle   	= 	{Proceedings of the Association for Machine Translation in the Americas},
    year        	= 	{2006},
    pages       	= 	{223--231}
}

@article{Xue2021mt5,
	author			=	{Linting Xue and Noah Constant and Adam Roberts and Mihir Kale and Rami Al-Rfou and Aditya Siddhant and Aditya Barua and Colin Raffel},
	title			=	{{mT5}: A massively multilingual pre-trained text-to-text transformer},
	journal			=	{arXiv preprint arXiv:2010.11934},
	year			=	{2021}
}

@article{martins2024eurollm,
  title             = {Eurollm: Multilingual language models for europe},
  author            = {Martins, Pedro Henrique and Fernandes, Patrick and Alves, Jo{\~a}o and Guerreiro, Nuno M and Rei, Ricardo and Alves, Duarte M and Pombal, Jos{\'e} and Farajian, Amin and Faysse, Manuel and Klimaszewski, Mateusz and others},
  journal           = {arXiv preprint arXiv:2409.16235},
  year              = {2024}
}

@article{Vidal2023bwer,
   title            = {End-to-End page-Level assessment of handwritten text recognition},
   volume           = {142},
   ISSN             = {0031-3203},
   journal          = {Pattern Recognition},
   author           = {Vidal, Enrique and Toselli, Alejandro H. and Ríos-Vila, Antonio and Calvo-Zaragoza, Jorge},
   year             = {2023},
   pages            = {109695}
}

@inproceedings{casacuberta2024findings,
  title             = {Findings of a Machine Translation Shared Task Focused on Covid-19 Related Documents},
  author            = {Casacuberta, Francisco and Ceausu, Alexandru and Choukri, Khalid and Deligiannis, Miltos and Domingo, Miguel and Garc{\'\i}a-Mart{\'\i}nez, Mercedes and Herranz, Manuel and Jacquet, Guillaume and Papavassiliou, Vassilis and Piperidis, Stelios and others},
  booktitle         = {Proceedings of the Annual Conference of the Spanish Association for Natural Language Processing},
  year              = {2024}
}

@inproceedings{barrault2020findings,
    title           = "Findings of the 2020 Conference on Machine Translation ({WMT}20)",
    author          = {Barrault, Lo{\"i}c  and Biesialska, Magdalena  and Bojar, Ond{\v{r}}ej  and Costa-juss{\`a}, Marta R.  and Federmann, Christian  and Graham, Yvette  and Grundkiewicz, Roman  and Haddow, Barry  and Huck, Matthias  and Joanis, Eric  and Kocmi, Tom  and Koehn, Philipp  and Lo, Chi-kiu  and Ljube{\v{s}}i{\'c}, Nikola  and Monz, Christof  and Morishita, Makoto  and Nagata, Masaaki  and Nakazawa, Toshiaki  and Pal, Santanu  and Post, Matt  and Zampieri, Marcos},
    booktitle       = "Proceedings of the Fifth Conference on Machine Translation",
    year            = "2020",
    pages           = "1--55",
}

@article{dietterich1998approximate,
  title             = {Approximate statistical tests for comparing supervised classification learning algorithms},
  author            = {Dietterich, Thomas G},
  journal           = {Neural computation},
  volume            = {10},
  number            = {7},
  pages             = {1895--1923},
  year              = {1998},
}

@article{ulmer2022deep,
  title             = {Deep-significance-easy and meaningful statistical significance testing in the age of neural networks},
  author            = {Ulmer, Dennis and Hardmeier, Christian and Frellsen, Jes},
  journal           = {arXiv preprint arXiv:2204.06815},
  year              = {2022}
}

@article{rainio2024evaluation,
  title             = {Evaluation metrics and statistical tests for machine learning},
  author            = {Rainio, Oona and Teuho, Jarmo and Kl{\'e}n, Riku},
  journal           = {Scientific Reports},
  volume            = {14},
  number            = {1},
  pages             = {6086},
  year              = {2024},
}

@inproceedings{popovic2015chrf,
    title           = "chr{F}: character n-gram {F}-score for automatic {MT} evaluation",
    author          = "Popovi{\'c}, Maja",
    booktitle       = "Proceedings of the Tenth Workshop on Statistical Machine Translation",
    year            = "2015",
    pages           = "392--395"
}

@inproceedings{stanojevic2014beer,
  title             = {Beer: Better evaluation as ranking},
  author            = {Stanojevi{\'c}, Milo{\v{s}} and Sima’an, Khalil},
  booktitle         = {Proceedings of the Ninth Workshop on Statistical Machine Translation},
  pages             = {414--419},
  year              = {2014}
}

@inproceedings{kocmi2024wmt,
    title           = "Findings of the {WMT}24 General Machine Translation Shared Task: The {LLM} Era Is Here but {MT} Is Not Solved Yet",
    author          = "Kocmi, Tom  and Avramidis, Eleftherios  and Bawden, Rachel  and Bojar, Ond{\v{r}}ej  and Dvorkovich, Anton  and Federmann, Christian  and Fishel, Mark  and Freitag, Markus  and Gowda, Thamme  and Grundkiewicz, Roman  and Haddow, Barry  and Karpinska, Marzena  and Koehn, Philipp  and Marie, Benjamin  and Monz, Christof  and Murray, Kenton  and Nagata, Masaaki  and Popel, Martin  and Popovi{\'c}, Maja  and Shmatova, Mariya  and Steingr{\'i}msson, Steinth{\'o}r  and Zouhar, Vil{\'e}m",
    booktitle = "Proceedings of the Ninth Conference on Machine Translation",
    year            = "2024",
    pages           = "1--46"
}

@article{sparck1972fscore,
  title             = {A statistical interpretation of term specificity and its application in retrieval},
  author            = {Sparck Jones, Karen},
  journal           = {Journal of documentation},
  volume            = {28},
  number            = {1},
  pages             = {11--21},
  year              = {1972},
}

@article{gao2025llm,
  title             = {Llm-based nlg evaluation: Current status and challenges},
  author            = {Gao, Mingqi and Hu, Xinyu and Yin, Xunjian and Ruan, Jie and Pu, Xiao and Wan, Xiaojun},
  journal           = {Computational Linguistics},
  pages             = {1--27},
  year              = {2025},
}

@article{zhang2019bertscore,
  title             = {Bertscore: Evaluating text generation with bert},
  author            = {Zhang, Tianyi and Kishore, Varsha and Wu, Felix and Weinberger, Kilian Q and Artzi, Yoav},
  journal           = {arXiv preprint arXiv:1904.09675},
  year              = {2019}
}

@article{rei2020comet,
  title             = {COMET: A neural framework for MT evaluation},
author              = {Rei, Ricardo and Stewart, Craig and Farinha, Ana C and Lavie, Alon},
  journal           = {arXiv preprint arXiv:2009.09025},
  year              = {2020}
}

@inproceedings{ali2022xai,
  title             = {XAI for transformers: Better explanations through conservative propagation},
  author            = {Ali, Ameen and Schnake, Thomas and Eberle, Oliver and Montavon, Gr{\'e}goire and M{\"u}ller, Klaus-Robert and Wolf, Lior},
  booktitle         = {International conference on machine learning},
  pages             = {435--451},
  year              = {2022},
}

@article{nguyen2023inspecting,
  title             = {Inspecting explainability of transformer models with additional statistical information},
  author            = {Nguyen, Hoang C and Lee, Haeil and Kim, Junmo},
  journal           = {arXiv preprint arXiv:2311.11378},
  year              = {2023}
}

@article{kashefi2023explainability,
  title             = {Explainability of vision transformers: A comprehensive review and new perspectives},
author              = {Kashefi, Rojina and Barekatain, Leili and Sabokrou, Mohammad and Aghaeipoor, Fatemeh},
  journal           = {arXiv preprint arXiv:2311.06786},
  year              = {2023}
}

@article{li2025probing,
  title             = {Probing out-of-distribution generalization in machine learning for materials},
  author            = {Li, Kangming and Rubungo, Andre Niyongabo and Lei, Xiangyun and Persaud, Daniel and Choudhary, Kamal and DeCost, Brian and Dieng, Adji Bousso and Hattrick-Simpers, Jason},
  journal           = {Communications Materials},
  volume            = {6},
  number            = {1},
  pages             = {9},
  year              = {2025},
}

@article{kim2024investigation,
title               = {Investigation of out-of-distribution detection across various models and training methodologies},
journal             = {Neural Networks},
volume              = {175},
pages               = {106288},
year                = {2024},
author              = {Byung Chun Kim and Byungro Kim and Yoonsuk Hyun},
}

@inproceedings{kudugunta2023madlad,
 author             = {Kudugunta, Sneha and Caswell, Isaac and Zhang, Biao and Garcia, Xavier and Xin, Derrick and Kusupati, Aditya and Stella, Romi and Bapna, Ankur and Firat, Orhan},
 booktitle          = {Advances in Neural Information Processing Systems},
 pages              = {67284--67296},
 title              = {MADLAD-400: A Multilingual And Document-Level Large Audited Dataset},
 volume             = {36},
 year               = {2023}
}

@InProceedings{Tiedemann2020opusmt,
  author            = {J{\"o}rg Tiedemann and Santhosh Thottingal},
  title             = {{OPUS-MT} — {B}uilding open translation services for the {W}orld},
  booktitle         = {Proceedings of the 22nd Annual Conferenec of the European Association for Machine Translation (EAMT)},
  year              = {2020},
 }

@article{liu2020mbart,
    title           = "Multilingual Denoising Pre-training for Neural Machine Translation",
    author          = "Liu, Yinhan  and Gu, Jiatao  and Goyal, Naman  and Li, Xian  and Edunov, Sergey  and Ghazvininejad, Marjan  and Lewis, Mike  and Zettlemoyer, Luke",
    journal         = "Transactions of the Association for Computational Linguistics",
    volume          = "8",
    year            = "2020",
    pages           = "726--742",
}

@article{costa2022nllb,
  title         = {No language left behind: Scaling human-centered machine translation},
  author        = {Costa-juss{\`a}, Marta R and Cross, James and {\c{C}}elebi, Onur and Elbayad, Maha and Heafield, Kenneth and Heffernan, Kevin and Kalbassi, Elahe and Lam, Janice and Licht, Daniel and Maillard, Jean and others},
  journal       = {arXiv preprint arXiv:2207.04672},
  year          = {2022}
}

@article{cheng2025seed,
  title         = {Seed-x: Building strong multilingual translation llm with 7b parameters},
  author        = {Cheng, Shanbo and Bao, Yu and Cao, Qian and Huang, Luyang and Kang, Liyan and Liu, Zhicheng and Lu, Yu and Zhu, Wenhao and Chen, Jingwen and Huang, Zhichao and others},
  journal       = {arXiv preprint arXiv:2507.13618},
  year          = {2025}
}

@article{fan2021m2m,
  title         = {Beyond english-centric multilingual machine translation},
  author        = {Fan, Angela and Bhosale, Shruti and Schwenk, Holger and Ma, Zhiyi and El-Kishky, Ahmed and Goyal, Siddharth and Baines, Mandeep and Celebi, Onur and Wenzek, Guillaume and Chaudhary, Vishrav and others},
  journal       = {Journal of Machine Learning Research},
  volume        = {22},
  number        = {107},
  pages         = {1--48},
  year          = {2021}
}

@inproceedings{morris2004and,
  title         = {From WER and RIL to MER and WIL: improved evaluation measures for connected speech recognition.},
  author        = {Morris, Andrew Cameron and Maier, Viktoria and Green, Phil D},
  booktitle     = {Interspeech},
  pages         = {2765--2768},
  year          = {2004}
}

@inproceedings{deldjoo2025toward,
author          = {Deldjoo, Yashar and Mehta, Nikhil and Sathiamoorthy, Maheswaran and Zhang, Shuai and Castells, Pablo and McAuley, Julian},
title           = {Toward Holistic Evaluation of Recommender Systems Powered by Generative Models},
year            = {2025},
booktitle       = {Proceedings of the 48th International ACM SIGIR Conference on Research and Development in Information Retrieval},
pages           = {3932–3942}
}

@article{haag2024development,
title           = {Development of the holistic quality model and assessment – Integrating the economic quality aspect and establishing an extended interrelation analysis},
journal         = {Developments in the Built Environment},
volume          = {19},
pages           = {100511},
year            = {2024},
author          = {Phillip Haag and Laura Balangé and Roberta {Di Bari} and Kathrin Braun and Julia Weißert and Li Zhang and Volker Schwieger and Philip Leistner and Cordula Kropp and Hans Christian Jünger},
}

@article{freitag2021experts,
  title         = {Experts, errors, and context: A large-scale study of human evaluation for machine translation},
  author        = {Freitag, Markus and Foster, George and Grangier, David and Ratnakar, Viresh and Tan, Qijun and Macherey, Wolfgang},
  journal       = {Transactions of the Association for Computational Linguistics},
  volume        = {9},
  pages         = {1460--1474},
  year          = {2021},
}

@article{wu2022survey,
title           = {A survey of human-in-the-loop for machine learning},
journal         = {Future Generation Computer Systems},
volume          = {135},
pages           = {364-381},
year            = {2022},
author          = {Xingjiao Wu and Luwei Xiao and Yixuan Sun and Junhang Zhang and Tianlong Ma and Liang He},
}

\end{document}